\documentclass{article}

\usepackage{arxiv}

\usepackage[utf8]{inputenc} 
\usepackage[T1]{fontenc}    
\usepackage{hyperref}       
\usepackage{url}            
\usepackage{booktabs}       
\usepackage{amsfonts}       
\usepackage{nicefrac}       
\usepackage{microtype}      
\usepackage{lipsum}		
\usepackage{graphicx}
\usepackage{natbib}
\usepackage{doi}
\usepackage{multirow}
\usepackage{subfigure}
\usepackage{amssymb}
\usepackage{amsmath}
\usepackage{setspace}

\pdfoutput=1
\hypersetup{
    colorlinks=true,
    linkcolor=black,
    citecolor=black,
    filecolor=black,
    urlcolor=black,
}

\title{Counting and Locating High-Density Objects Using Convolutional Neural Network}

\date{February, 2021}	

\author{ \href{https://orcid.org/0000-0002-2273-8968}{\includegraphics[scale=0.06]{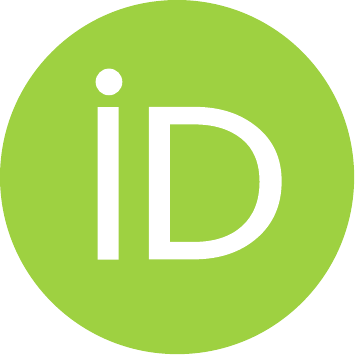}\hspace{1mm}Mauro dos Santos de Arruda} \\
	Federal University of Mato Grosso do Sul\\
	Campo Grande, MS, Brazil \\
	\texttt{mauro.arruda@ufms.br} \\
	\And
    \href{https://orcid.org/0000-0002-0258-536X}{\includegraphics[scale=0.06]{orcid.pdf}\hspace{1mm}Lucas Prado Osco}\\
	University of Western São Paulo\\
	Presidente Prudente, SP, Brazil \\
	\texttt{lucasosco@unoeste.br} \\
	\And
	\href{https://orcid.org/0000-0000-0000-0000}{\includegraphics[scale=0.06]{orcid.pdf}\hspace{1mm}Plabiany Rodrigo Acosta} \\
	Federal University of Mato Grosso do Sul\\
	Campo Grande, MS, Brazil \\
	\texttt{plabiany@gmail.com} \\
	\And
	\href{https://orcid.org/0000-0002-4527-5724}{\includegraphics[scale=0.06]{orcid.pdf}\hspace{1mm}Diogo Nunes Gon\c{c}alves}\\
	Federal University of Mato Grosso do Sul\\
	Campo Grande, MS, Brazil \\
	\texttt{dnunesgoncalves@gmail.com} \\
	\And
	\href{https://orcid.org/0000-0002-9096-6866}{\includegraphics[scale=0.06]{orcid.pdf}\hspace{1mm}José Marcato Junior} \\
	Federal University of Mato Grosso do Sul\\
	Campo Grande, MS, Brazil \\
	\texttt{jose.marcato@ufms.br} \\
	\And
	\href{https://orcid.org/0000-0001-6633-2903}{\includegraphics[scale=0.06]{orcid.pdf}\hspace{1mm}Ana Paula Marques Ramos} \\
	University of Western São Paulo\\
	Presidente Prudente, SP, Brazil \\
	\texttt{anaramos@unoeste.br} \\
	\And
	\href{https://orcid.org/0000-0002-4471-0886}{\includegraphics[scale=0.06]{orcid.pdf}\hspace{1mm}Edson Takashi Matsubara} \\
	Federal University of Mato Grosso do Sul\\
	Campo Grande, MS, Brazil \\
	\texttt{edsontm@facom.ufms.br} \\
	\And
	\href{https://orcid.org/0000-0000-0000-0000}{\includegraphics[scale=0.06]{orcid.pdf}\hspace{1mm}Zhipeng Luo} \\
	Xiamen University\\
	Xiamen, FJ, China \\
	\texttt{zpluo@stu.xmu.edu.cn} \\
	\And
	\href{https://orcid.org/0000-0001-7899-0049}{\includegraphics[scale=0.06]{orcid.pdf}\hspace{1mm}Jonathan Li} \\
	University of Waterloo\\
	Waterloo, ON, Canada \\
	\texttt{junli@uwaterloo.ca} \\
	\And
	\href{https://orcid.org/0000-0002-8274-2707}{\includegraphics[scale=0.06]{orcid.pdf}\hspace{1mm}Jonathan de Andrade Silva} \\
	Federal University of Mato Grosso do Sul\\
	Campo Grande, MS, Brazil \\
	\texttt{jonathan.andrade@ufms.br} \\
	\And
	\href{https://orcid.org/0000-0002-8815-6653}{\includegraphics[scale=0.06]{orcid.pdf}\hspace{1mm}Wesley Nunes Gonçalves}\thanks{corresponding author: wesley.goncalves@ufms.br} \\
	Federal University of Mato Grosso do Sul\\
	Campo Grande, MS, Brazil \\
	\texttt{wesley.goncalves@ufms.br} \\
}

\renewcommand{\shorttitle}{\textit{arXiv - Arruda et al. (2021)}}

\hypersetup{
pdftitle={A Deep Learning Approach Based on Graphs to Detect Plantation Lines},
pdfsubject={cs.CV},
pdfauthor={Gonçalves, L.P. et al.},
pdfkeywords={remote sensing, convolutional neural networks, aerial imagery, UAV, object detection},
}

\begin{document}
\maketitle

\begin{abstract}
This paper presents a Convolutional Neural Network (CNN) approach for counting and locating objects in high-density imagery. To the best of our knowledge, this is the first object counting and locating method based on a feature map enhancement and a Multi-Stage Refinement of the confidence map. The proposed method was evaluated in two counting datasets: tree and car. For the tree dataset, our method returned a mean absolute error (MAE) of 2.05, a root-mean-squared error (RMSE) of 2.87 and a coefficient of determination (R$^2$) of 0.986. For the car dataset (CARPK and PUCPR+), our method was superior to state-of-the-art methods. In the these datasets, our approach achieved an MAE of 4.45 and 3.16, an RMSE of 6.18 and 4.39, and an R$^2$ of 0.975 and 0.999, respectively. The proposed method is suitable for dealing with high object-density, returning a state-of-the-art performance for counting and locating objects.
\end{abstract}


\section{Introduction}
\label{sec:introduction}

In computer vision, counting and locating objects in images have been the attention of several approaches \citep{SINDAGI20183}. These methods help to control and count people \citep{Idrees2018ECCV}, support car detection \citep{Hsieh2017}, monitor wildlife \citep{arruda2018IJCNN}, and even count bacterial colonies \citep{FERRARI2017629}. As expected, the majority of these methods are based on the well-known object detection task, including the recent methods based on convolutional neural networks (Faster R-CNN \citep{Ren2017}, Mask-RCNN \citep{He2017}, RetinaNet \citep{Lin2020}), multi-scale variants (Multi-Scale Structures \citep{OHNBAR2017557}, Multi-scale deep feature learning network \citep{MA2020107149}, Gated CNN \citep{YUAN2019107131}) and ensembles of models \citep{XU2020107098}. Many object detection methods consider a bounding box (bbox) around the targeted objects and can provide both location (center of the bbox) and counting (number of bboxs). Recent contribution to this matter is from \cite{Hsieh2017}, where the authors proposed, simultaneously, Layout Proposal Networks (LPNs) and spatial kernels to detect objects in video. These additions helped to improve the object counting and location using an object detection framework. Still, even state-of-the-art methods return bounding boxes that partially overlap multiple objects, which is still a problem since the adjacent object region is detected as a separate object \citep{Goldman2019CVPR}.

One of the biggest challenges regarding the counting and location of objects in images is the high-object density. Object detection methods are, in general, not adequate for high-density scenes \citep{Goldman2019CVPR}. In this scenario, overlapping objects are difficult to analyze due to the size of the instances and the standpoint of the scene. Thus, approaches that model the problem of counting objects with a density estimation has been defined as state-of-the-art solutions, and are providing interesting solutions for dense scenes \citep{Goldman2019CVPR,Aich2018ImprovingOC}. In \cite{Goldman2019CVPR}, the authors proposed a CNN-based detection method, using the bounding box, to cope with densely packed scenes. They considered a layer to estimate a quality score index and used a novel EM (Expectation-Maximization) merging unit to solve the overlap ambiguities with this score. However, handling high-density objects in images is still a concerning issue, both in counting and locating objects.

Another problem regarding object count from detection frameworks is the need of detailed ground-truth labeled data, which is hard to obtain at large-scales \citep{ILSVRC15}. Acquiring a large-scale annotated data is a time-consuming process. Because of that, the approach based on a lighter weight image label is something that researchers have previously proposed \citep{ZHANG201868,Fiaschi2012}. Still, recent studies are implementing point annotations to reduce the supervision task \citep{Aich2018ImprovingOC,Liu2019a}. Point annotations are easier to obtain than bounding-boxes, and many counting and locating approaches do not need to rely on them to identify an object \citep{Liu2019a}. These types of approaches can rely on context information, and, for most problems, object instances will share a similar color, texture, and shape; meaning that the method will learn how to recognize them even if only using point features \citep{Aich2018ImprovingOC}.

Recently, state-of-the-art methods to count objects include the VGG-GAP and VGG-GAP-HR \citep{Aich2018ImprovingOC} approaches, Layout Proposal Networks (LPN) \citep{Hsieh2017} and Deep IoU CNN \citep{Goldman2019CVPR}. These methods were applied in counting and locating cars, crowds, biological cells and products from supermarket shelves, returning impressive performances in high-density scenes. Despite the promising results, scale variations, clutter background, occlusions, and especially high-density of objects are still challenges that hinder methods of providing high-quality predictions. That way, in previous work, we developed an initial model for the location and counting of Citrus-trees in UAV multispectral images \citep{OSCO202097}. This initial model significantly surpassed methods for detecting objects such as RetinaNet and Faster-RCNN.

In this paper, we propose a method for counting and locating objects based on convolutional neural network \citep{Simonyan14c}. Unlike previous research that has estimated a rectangle for each object, the present method is based on a density estimation map with the confidence that an object occurs in each pixel, following \citep{Aich2018ImprovingOC}. Different from previous work \citep{OSCO202097}, the proposed method uses a feature map enhancement with a Pyramid Pooling Module (PPM) \citep{Zhao2017CVPR} that allows our method to incorporate global information at different scales. Consequently, the proposed method incorporates sufficient global context information for a good characterization of objects similarly to \cite{ZHANG2019166} with its hierarchical context module. Thus, we hypothesize that this approach is best suitable for situations of high object density. 

Aside from that, another potential pitfall of previous methods is the missed detections due to object occlusion and high-density. To compensate for these detections, and produce the correct count label, a Multi-Stage Refinement over the ground-truth map is proposed. By implementing multiple stages, our method provides hierarchical learning of the object position, starting from a rough to a more refined prediction of the center of the object. Therefore, our method is divided into four main phases: 1) a feature map generation using a CNN; 2) feature map enhancement with a Pyramid Pooling Module (PPM); 3) a Multi-Stage Refinement of the confidence maps; and 4) the obtention of the object position through peaks in the confidence map.

Experiments were performed in three datasets. First, we evaluated the method parameters in a tree counting dataset containing 3,370 images and aproximately 232,000 objects. Second, we verify its generalization by comparing it with state-of-the-art methods in two car-counting benchmarks: CARPK and PUCPR+. The proposed method outperforms state-of-the-art methods for counting purposes in these three benchmarks. To the best of our knowledge, this is the first object counting and locating method based on a feature map enhancement and a multi-stage refinement of the confidence map.

\section{Proposed Method}
\label{sec:proposedApproach}

This section describes our method to count and locate objects. This method uses a three-channel image, with $w \times h$ pixels as input and processes it with a CNN. The object counting and location were modeled after a 2D confidence map estimation, following the procedures presented in \citep{Aich2018ImprovingOC}.

The confidence map is a 2D representation of the likelihood of an object occurring in each pixel. 
We improved the confidence map estimation by including global and local information through a Pyramid Pooling Module (PPM) \citep{Zhao2017CVPR}. We also proposed a multi-stage prediction phase to refine the confidence map to a more accurate prediction of the center of the objects.

\begin{figure*}[ht]
	\centering
	\includegraphics[width=1\textwidth]{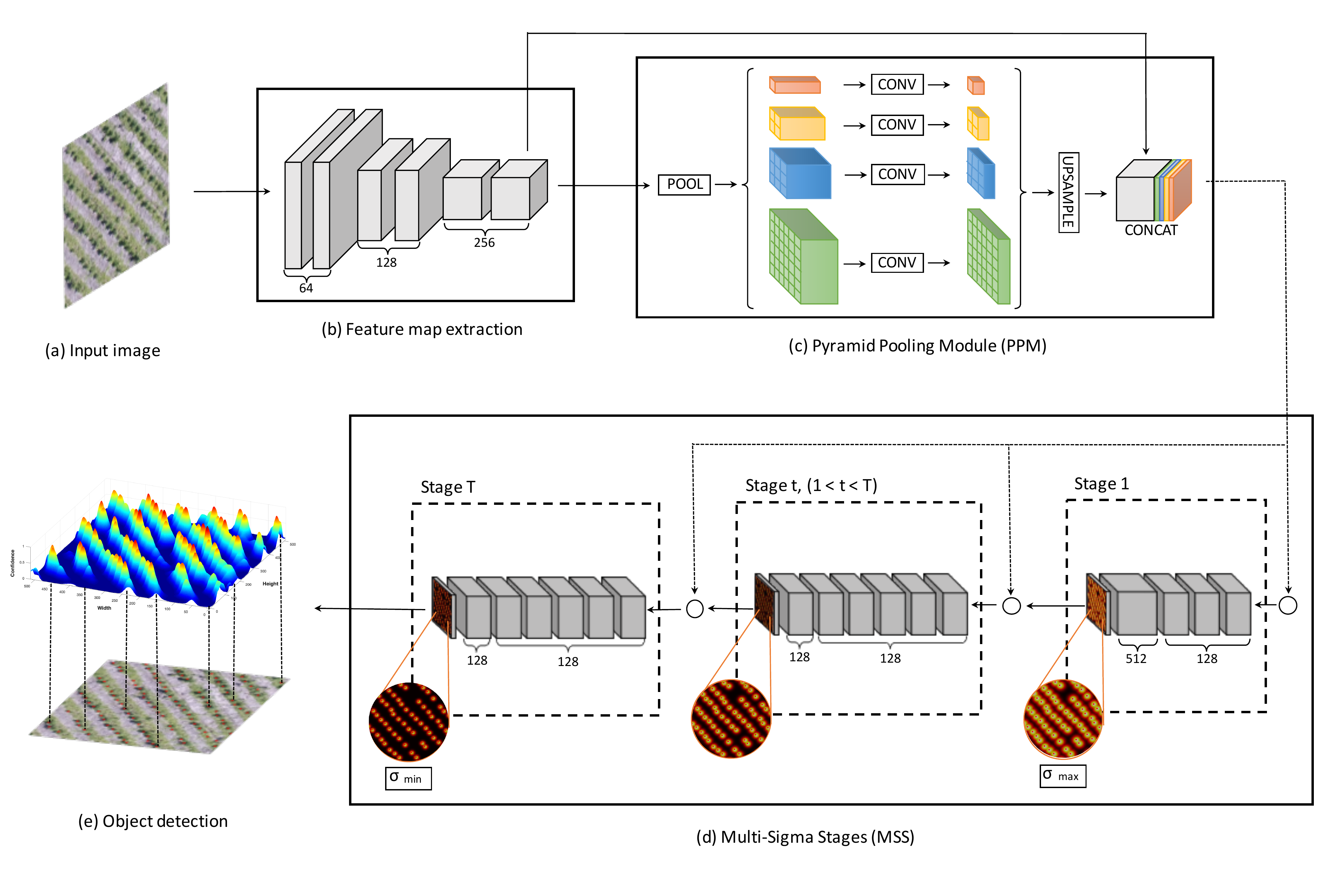}
	\caption{Our method for the confidence map prediction using the Pyramid Pooling Module (PPM) and the multi-stage refinement approach. The initial part (b), based on VGG19 \citep{Simonyan14c}, extracts a feature map from the input image (a). This feature map is used as input for the PPM (c) \citep{Zhao2017CVPR}. The resulting volume is then used as input to the first stage of a Multi-Sigma Stages (MSS) phase (d) \citep{Aich2018ImprovingOC}. The concatenation of the PPM and the prediction map of the previous stage is used as input for the remaining stages. The $T$ stages apply a standard deviation ($\sigma$) for the confidence map peak, starting at maximum-to-minimum so that values are spaced equally.}
	\label{fig:proposta}
\end{figure*}

Fig. \ref{fig:proposta} illustrates the phases of the proposed method, which are detailed in the following section. Our approach is divided into four main phases: 1) feature map generation with a CNN (Section \ref{subsec:feature_map}); 2) feature map enhancement with the PPM (Section \ref{subsec:PSPModule}); 3) multi-stage refinement of the confidence map (Sections  \ref{subsec:multiSigma} and \ref{subsec:generateMap}); and, 4) object position obtention by peaks in the confidence map (Section \ref{subsec:plantLocalization}).

\subsection{Feature Map Using CNN}
\label{subsec:feature_map}

We started by extracting a feature map with a CNN from a given input image (Fig. \ref{fig:proposta}(a)).
This feature map extraction module is based on the VGG19 \citep{Simonyan14c}, where the first two convolutional layers have 64 filters of a $3 \times 3$ size and are followed by a maximum pooling layer with a $2 \times 2$ window. The last two convolutional layers have 256 filters with a $3 \times 3$ size. All convolutional layers use the rectified linear unit (ReLU) function, with a stride of 1 and zero-padding, returning an output with the same resolution as the input.

We evaluated two variations of our method for different input images dimensions. The first variation receives an input image with $512 \times 512$ resolution and produces a feature map in the final layer with $64 \times 64$ resolution. Proportionally, the second variation receives an input image with $1024 \times 1024$ pixels, and the output feature map has a resolution of $128 \times 128$. Despite the low resolution, this map can describe the relevant features extracted from the image.

\subsection{Improving Feature Map with Pyramid Pooling Module}
\label{subsec:PSPModule}

Many CNN cannot incorporate sufficient global context information to ensure a good performance in characterizing high-density objects. To solve this issue, our method adopts a global and subregional context module called PPM \citep{Zhao2017CVPR}. This module allows CNN to be invariant to scale. Fig. \ref{fig:proposta} (c) illustrates the PPM that combines the features of four pyramid scales, with resolutions of $1 \times 1$, $2 \times 2$, $3 \times 3$ and $6 \times 6$, respectively. The highest general level, displayed in the color orange, applies a global max pooling which creates a $1 \times1$ feature map to describe the global image context, such as the number of detected objects in the image. The other levels divide the input map into subregions, forming a grouped representation of the image with their subcontext informations. 

The levels of the PPM contain feature maps with various sizes. Because of this, we used a $1 \times 1$ convolution layer with 512 filters after each level. We upsampled the feature maps to the same size as the input map with bilinear interpolation. Lastly, these feature maps are concatenated with the input map to form an improved description of the image. This step ensures that small object information is not lost in the PPM phase. Although this module is proposed for semantic segmentation, it has proven to be a robust method for counting objects according to our experiments, as it includes information at different scales and global context.

\subsection{Multi-Stage Refinement}
\label{subsec:multiSigma}

In the multi-stage refinement phase, the improved feature map obtained by PPM is used as input for the $T$ stages that estimates the confidence map. The first stage (Fig. \ref{fig:proposta} (d)) receives the feature map and generates the confidence map $C_1$ by using five convolutional layers: three layers with 128 filters with a $3 \times 3$ size; one layer with 512 filters with a $1 \times 1$ size; and one layer with a single filter, corresponding to the confidence map.

At a subsequent stage $t$ (Fig. \ref{fig:proposta} (d)), the prediction returned by the previous stage $C_{t-1}$ and the feature map from the PPM process are concatenated. They are used to produce a refined confidence map $C_t$. The $T-1$ final stages consist of seven convolutional layers: five layers with 128 filters with a $7 \times 7$ size; and one layers with 128 filters with a $1 \times 1$ size. The last layers have a sigmoid activation function so that each pixel represents the probability of the occurrence of an object (values between $[0,1]$). The remaining layers have a ReLU activation function. Through multiple stages, we proposed hierarchical learning of the center of the object. The first stage roughly predicts the position, while the other stages refine this prediction (Fig. \ref{fig:activationStages}).

To avoid the vanishing gradient problem during the training phase, we adopted a loss function (Eq. \ref{eq:loss}) to be applied at the end of each stage.

\begin{equation}
	f_t = \sum_p \parallel \hat{C_t}(p) - C_{t}(p)\parallel^2_2,
	\label{eq:loss}
\end{equation}
where $\hat{C_t}$ is the ground truth confidence map of the stage $t$ (Section \ref{subsec:generateMap}). The overall loss function is given by:

\begin{equation}
	f = \sum_{t=1}^{T} f_t
\end{equation}

\subsection{Generation of Confidence Maps}
\label{subsec:generateMap}

\begin{figure}[ht]
	\centering
	\subfigure[RGB images]{\includegraphics[width=.24\columnwidth]{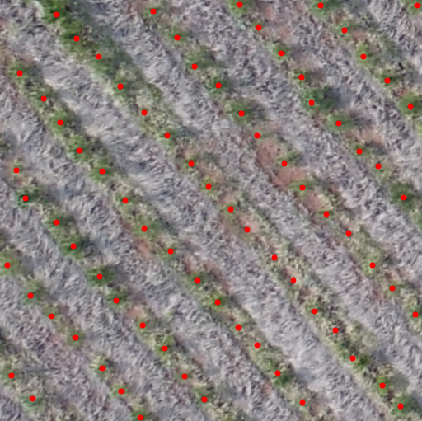}}
	\subfigure[$\sigma_t = 1.5$]{\includegraphics[width=.24\columnwidth]{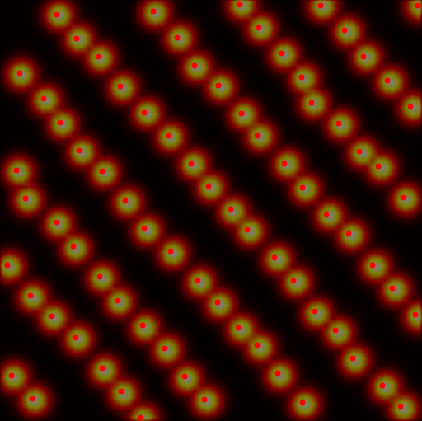}}
	\subfigure[$\sigma_t = 1.0$]{\includegraphics[width=.24\columnwidth]{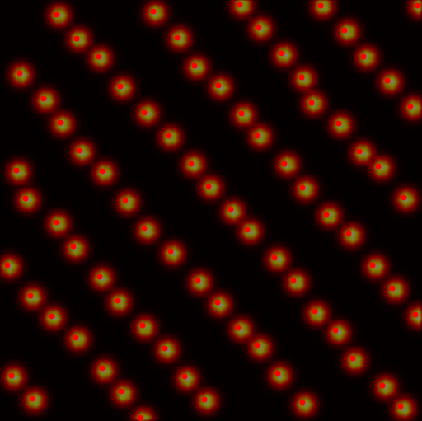}}
	\subfigure[$\sigma_t = 0.5$]{\label{fig:sigmab}\includegraphics[width=.24\columnwidth]{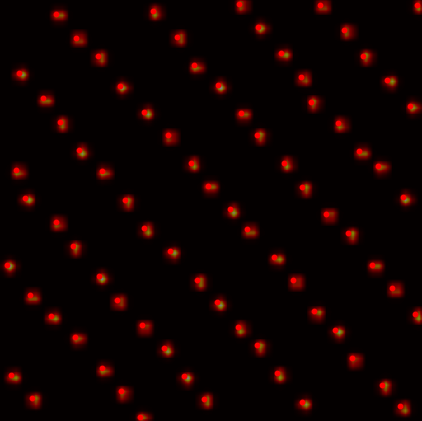}}
	
	\caption{\label{fig:sigma} Example of an RGB image and its corresponding ground-truth confidence maps with different $\sigma_t$ values.}
\end{figure}

As mentioned in the previous section, to train our method, a confidence map $\hat{C_t}$ is generated as a ground truth for each stage $t$ by using the center of the objects as annotations in the image. The $\hat{C_t}$  is generated by placing a 2D Gaussian kernel at each center of the labeled objects \citep{Aich2018ImprovingOC}. The Gaussian kernel has a standard deviation ($\sigma_t$) that controls the spread of the confidence map peak, as shown in Fig. \ref{fig:sigma}..

Our approach uses different values of $\sigma_t$ for each stage $t$ to refine the object center prediction during each stage. The $\sigma_1$ of the first stage is set to a maximum value ($\sigma_{max}$) while the $\sigma_T$ of the last stage is set to a minimum value ($\sigma_{min}$). The appropriate values of $\sigma_{max}$ and $\sigma_{min}$ are evaluated in the experiments. The $\sigma_t$ for each intermediate stage is equally spaced between $[\sigma_{max}, \sigma_{min}]$. The early stage should return a rough prediction of the center of the object, and this prediction is refined in the subsequent stages.

Fig. \ref{fig:sigma} illustrates an example of a ground truth confidence map with three values of $\sigma_t$. Fig. \ref{fig:sigma} (a) shows the RGB image and the locations of each objected marked by a red dot. Fig.\ref{fig:sigma} (b, c, and d) present the ground truth confidence maps for $\sigma_t = 0.5, 1.0$ and $1.5$, respectively.  In our experiment, the usage of different $\sigma$ helped refine the confidence map, improving its robustness.

\subsection{Object Localization with Confidence Map}
\label{subsec:plantLocalization}

\begin{figure}[ht]
	\centering
	\setcounter{subfigure}{0}
	\subfigure{\includegraphics[width=.8\columnwidth]{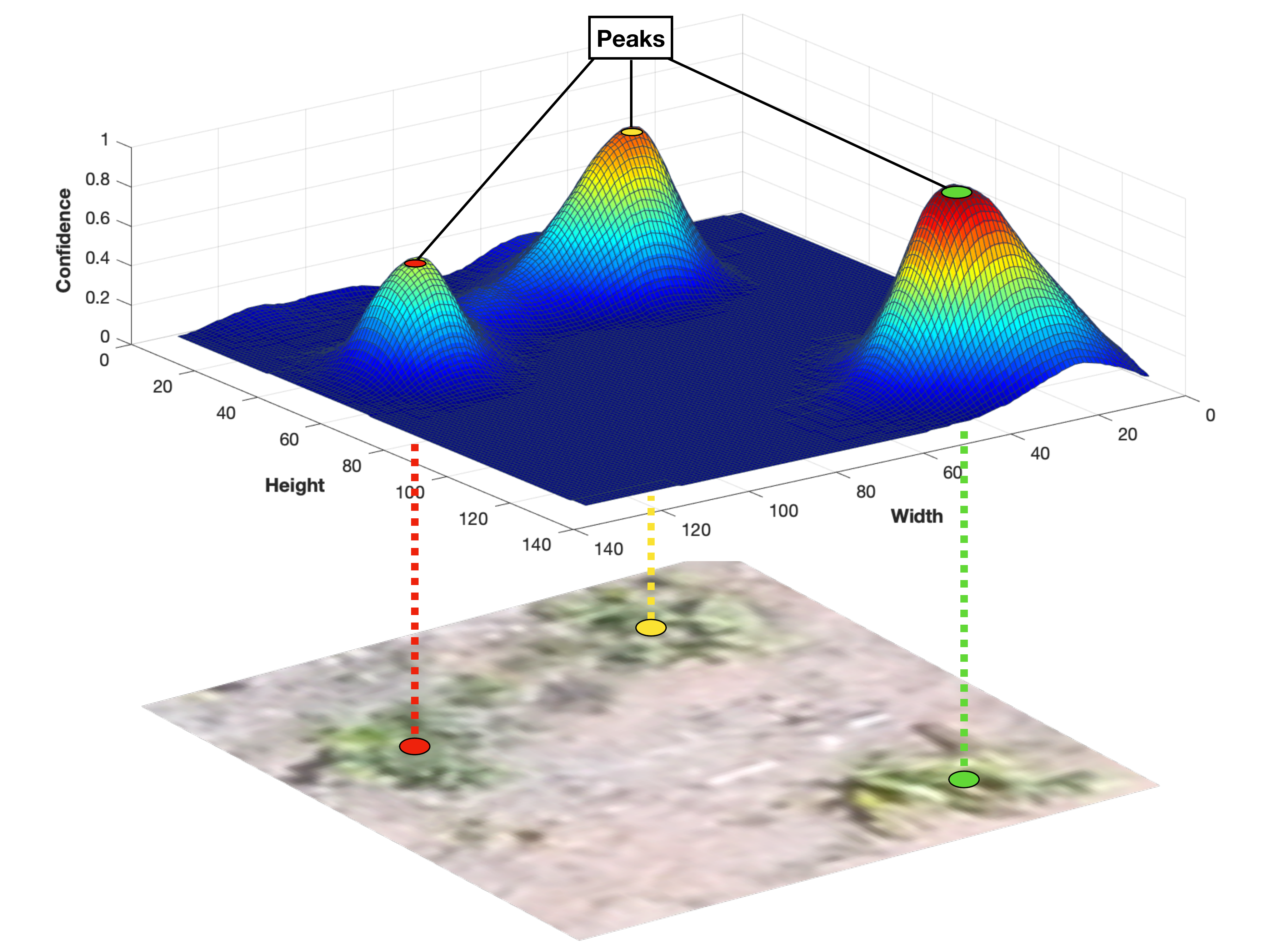}}
	
	\caption{\label{fig:heatMapPicos} Example of the localization of eucalyptus trees from a refined confidence map.}
\end{figure}

Object locations are obtained from the confidence map of the last stage ($C_T$). We estimate the peaks (local maximum) of the confidence map by analyzing the 4-pixel neighborhood of each given location of $p$. Thus, $p = (x_p, y_p)$ is a local maximum if $C_{T} (p) > C_{T} (v)$ for all the neighbors $v$, where $v$ is given by $(x_p \pm 1, y_p)$ or $(x_p, y_p \pm 1)$. An example of the object location from the confidence map peaks is shown in Fig. \ref{fig:heatMapPicos} .

To avoid noise or low probability of occurrence of the position $p$, a peak in the confidence map is considered as an object only if $C_ {T} (p) > \tau$. Besides that, we set a minimum distance $\delta$ to allow the method to detect very close objects. After a preliminary experiment, we used $\tau = 0.35$ and $\delta = 1$ pixel that allows the detection of objects from two pixels of distances.

\section{Experiments}

\subsection{Image Datasets}
\label{subsec:imageCollection}

\begin{figure}[ht]
	\centering
	
	\subfigure{\includegraphics[width=.32\columnwidth]{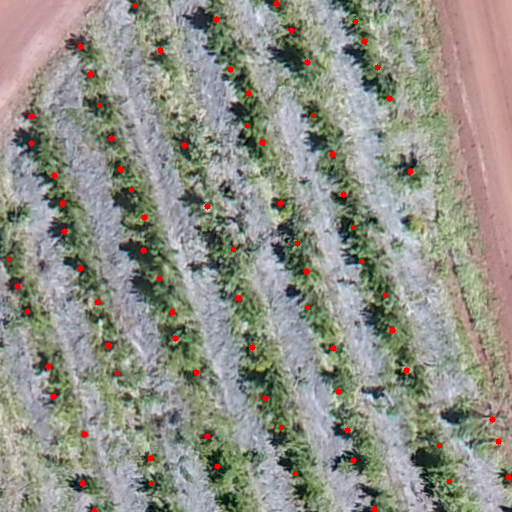}}
	\setcounter{subfigure}{0}
	\subfigure{\includegraphics[width=.32\columnwidth]{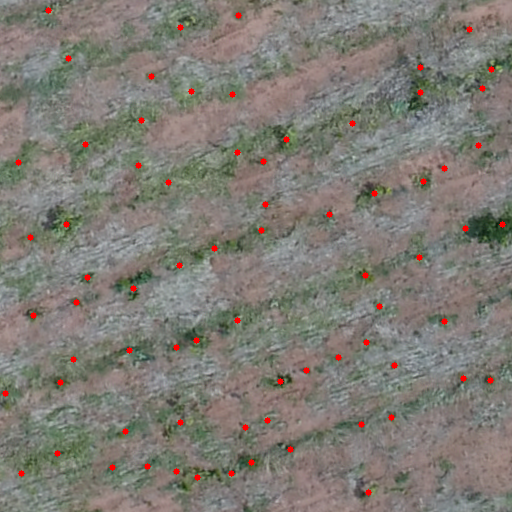}}
	\subfigure{\includegraphics[width=.32\columnwidth]{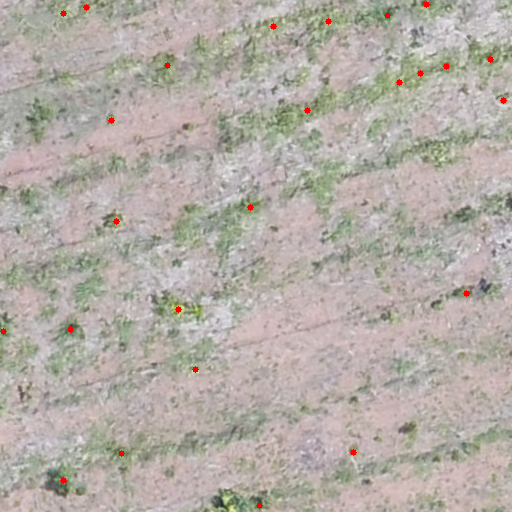}}
	
	\caption{Examples of the tree dataset. The eucalyptus trees are at different growth stages and plantation densities.}
	\label{fig:density}
\end{figure}

To test the robustness of our method, we evaluated it in a new and challenging dataset of eucalyptus tree images. We used this image dataset because there are different tree plantation densities, ranging from extreme cases to more sparse trees (Fig. \ref{fig:density}). This variation in density is a challenge for counting and locating objects. The trees were also at different growth stages. This permitted to evaluate the proposed method in different scales (tree size) and changes in appearance. 

The images were captured by an Unmanned Aerial Vehicle (UAV) in a rural property in Mato Grosso do Sul, Brazil, over four different areas of approximately 40 ha each. The eucalyptus trees were planted at different spacing, the densest being at 1.25 meters from each other, with an average of 1,750 trees per hectare. These trees were at different growth stages, variating between high and canopy areas. The images were acquired with an RGB sensor, which produced a pixel size of 4.15 cm. A total of four orthomosaic were generated from the area of interest. Approximately 232,000 eucalyptus trees were labeled as a point feature by a specialist.

To evaluate the robustness and generability of the proposed approach, we also compared the performance of our method in two well-known image datasets for counting cars: CARPK and PUCPR+ benchmarks \citep{Hsieh2017}. We compare the prediction metrics with state-of-the-art methods such One-Look Regression \citep{Mundhenk2016}, IEP Counting \citep{Stahl2019}, YOLO \citep{Redmon2017}, YOLO9000 \citep{Redmon2017}, Faster R-CNN \citep{Ren2017}, RetinaNet \citep{Lin2020,Hsieh2017} , LPN \citep{Hsieh2017}, VGG-GAP \citep{Aich2018ImprovingOC}, VGG-GAP-HR  \citep{Aich2018ImprovingOC} and Deep IoU CNN \citep{Goldman2019CVPR}.

\subsection{Experimental Setup}
\label{sec:setup}

The four orthomosaics were split into 3,370 patches with $512 \times 512$ pixels without overlapping. These patches were randomly divided into training ($n=2,870$), validation ($n=250$) and testing ($n=250$) sets. For training the CNN, we applied a Stochastic Gradient Descent optimizer with a momentum of 0.9. To reduce the risk of overfitting, we used the validation set for the hyperparameter tuning on the learning rate and the number of epochs. After minimal hyperparameter tuning, the learning rate was 0.01 and the number of epochs was equal to 100. Instead of training the proposed approach from scratch, we initialized the weights of the first part with pre-trained weights in ImageNet. Six regression metrics, the mean absolute error (MAE) \citep{wackerly2014mathematical, RMSEMAE2014}, root mean squared error (RMSE) \citep{wackerly2014mathematical, RMSEMAE2014}, the coefficient of determination (R$^2$) \citep{draper1998applied}, the Precision, Recall, and the F-Measure, were used to measure the performance.
Training and testing were perfomed in a deskop computer with Intel(R) Xeon(R) CPU E3-1270@3.80GHz, 64 GB memory, and NVIDIA Titan V Graphics Card (5120 Compute Unied Device Architecture - CUDA cores and 12 GB graphics memory). The methods were implemented using Keras-Tensorflow on the Ubuntu 18.04 operating system.

\section{Results and Discussion}
\label{sec:results}

This section presents and discusses the results obtained by the proposed method while comparing it with state-of-the-art methods. First, we demonstrate the influence of different parameters, which includes the $\sigma$ to generate the ground truth confidence maps, the number of stages necessary to refine the prediction, and the usage of PPM \citep{Zhao2017CVPR}  to include context information based on multiple scales. Second, we compare the results with the baseline of the proposed method. For this, we used the tree counting dataset and the car counting datasets (CARPK and PUCPR+).

\subsection{Parameter Analysis}
\label{subsubsec:ApproachParameters}

We present the results of the proposed method in the validation set for a different number of stages on the tree counting dataset. These stages are responsible for refining the confidence map. We observed that by using two stages ($T = 2$), the proposed method already returned satisfactory results (Table \ref{tab:stages}). When increasing to $T = 4$ stages, we obtained the best result, with MAE, RMSE, R$^2$, Precision, Recall, and F-Measure of 2.69, 3.57, 0.977, 0.817, 0.831, and 0.823, respectively. These results indicate the multi-stage refinement can affect the object counting tasks significantly. This is because the confidence map is refined in later stages, increasing the chance of objects be detected in high-density regions.

\begin{table}[ht]
	\renewcommand{\arraystretch}{1.3}
	\caption{Evaluation of the number of stages ($T$) on the validation set of the tree counting dataset using $\sigma_{min} = 1$ and $\sigma_{max} = 3$.}
	\small
	\centering
	\begin{tabular}{|c||c||c||c||c||c||c|}
		\hline
		Stages ($T$) & MAE & RMSE & R$^2$ &  Precision & Recall & F-Measure  \\
		\hline
		2 & 2.86 & 3.82 & 0.974 & 0.809 & 0.825 & 0.816 \\
		\textbf{4} & \textbf{2.69} & \textbf{3.57} & \textbf{0.977} & \textbf{0.817}  & \textbf{0.831} & \textbf{0.823} \\
		6 & 3.48 & 4.61 & 0.962 & 0.805 & 0.836 & 0.819 \\
		8 & 2.90 & 3.79 & 0.974 & 0.816 & 0.823 & 0.818 \\
		10 & 3.32 & 4.25 & 0.967 & 0.789 & 0.796 & 0.790 \\
		\hline
	\end{tabular}
	\label{tab:stages}
\end{table}

We evaluated the $\sigma_{min}$ and $\sigma_{max}$ responsible for generating the ground truth confidence maps implemented in the $T$ stages. In this experiment, we adopt $T = 4$ stages that achieved the best results from the previous experiment. The confidence map of the first stage is generated using $\sigma_{max}$, while the last stage uses $\sigma_{min}$, and the intermediate stages are constructed from values equally spaced between $[\sigma_{max}, \sigma_{min}]$. A low $\sigma$, relative to the object area (e.g., tree canopy) provides a confidence map without correctly covering the object's area. However, a high $\sigma$ generates a confidence map that, while fully covers the object, may include nearby objects in high-density conditions. These conditions make it difficult to spatially locate objects in the image.

The evaluation for $\sigma_{max}$ is presented in Table \ref{tab:maxSigma}. The highest result was obtained with $\sigma_{max} = 3$, which best covers the tree-canopies without overlapping them. Still, we observed that other values for $\sigma_{max}$ also returned good results. Since $\sigma_{max}$ is used in the first stage, it does a small influence over the final result, since the confidence map is refined in the subsequent stages.

\begin{table}[ht]
	\renewcommand{\arraystretch}{1.3}
	\caption{Evaluation of the $\sigma_{max}$ in the validation set of the tree counting dataset. We adopted the $\sigma_{min} = 1$ and stages $T = 4$.}
	\small
	\centering
	\begin{tabular}{|c||c||c||c||c||c||c|}
		\hline
		$\sigma_{max}$ & MAE & RMSE & R$^2$ & Precision & Recall & F-Measure  \\
		\hline
		2 & 3.31 & 4.31 & 0.966 & 0.811 & 0.837 & 0.822 \\
		\textbf{3} & \textbf{2.69} & \textbf{3.57} & \textbf{0.977} & \textbf{0.817}  & \textbf{0.831} & \textbf{0.823} \\
		4 & 3.21 & 4.24 & 0.968 & 0.804 & 0.816 & 0.809 \\
		\hline
	\end{tabular}
	\label{tab:maxSigma}
\end{table}

The results for the $\sigma_{min}$ are summarized in Table \ref{tab:minSigma}. The $\sigma_{min}$ has great influence over the final result since it is responsible for the last confidence map. The overall best result was obtained with a $\sigma_{min} = 1.0$, which achieved a MAE, RMSE, R$^2$, Precision, Recall and F-Measure of $2.69$, $3.57$, $0.977$, $0.817$, $0.831$ and $0.823$, respectively. This shows that the $\sigma_{min} = 1.0$ is the best fit for the size of the tree canopy. The conducted experiments showed that, with appropriate values of $\sigma_{max} = 3$ and $\sigma_{min} = 1$, high performance for counting trees can be obtained (Table \ref{tab:minSigma}).

\begin{table}[ht]
	\renewcommand{\arraystretch}{1.3}
	\caption{Evaluation of the $\sigma_{min}$ in the validation set of the tree counting dataset. We used $\sigma_{max} = 3$ and stages $T = 4$.}
	\small
	\centering
	\begin{tabular}{|c||c||c||c||c||c||c|}
		\hline
		\textbf{$\sigma_{min}$} & MAE	& RMSE & R$^2$ & Precision & Recall & F-Measure \\
		\hline
		0.5 & 11.01 & 13.77 & 0.658 & 0.868 & 0.721 & 0.783 \\
		0.75 & 2.93 & 3.89 & 0.972 & 0.820 & 0.831 & 0.824 \\
		\textbf{1} & \textbf{2.69} & \textbf{3.57} & \textbf{0.977} & \textbf{0.817}  & \textbf{0.831} & \textbf{0.823} \\
		1.25 & 3.05 & 4.01 & 0.970 & 0.815 & 0.822 & 0.817 \\
		1.5 & 2.94 & 3.73 & 0.975 & 0.818 & 0.810 & 0.813 \\
		\hline
	\end{tabular}
	\label{tab:minSigma}
\end{table}

To verify the potential of our method in real-time processing, we perform a comparison of the processing time performance for different amounts of stages ($T$). Table \ref{tab:timeApproach} shows the processing time of the proposed method for values of $T$ = $2$, $4$, $6$, $8$ and $10$. For this, we used 100 images from the tree test set and extracted the average processing time and standard deviation. We used the values of $\sigma_{min} = 1$ and $\sigma_{max} = 3$ that obtained the best performance in the previous tests. The results showed that the proposed approach can achieve real-time processing. For the best configuration with stages $T = 4$ the approach can deliver an image detection in $1.42$ seconds with a standard deviation of $0.028$. 

\begin{table}[ht]
	\renewcommand{\arraystretch}{1.3}
	\caption{Processing time evaluation of the proposed approach for different amounts of $T$.}
	\small
	\centering
	\begin{tabular}{|c||c||c|}
		\hline
		Stages ($T$) & Average Time (s) & Standard deviation \\
		\hline
		2 & 0.802 & 0.022 \\
		4 & 1.426 & 0.028 \\
		6 & 2.063 & 0.058  \\
		8 & 2.675 & 0.059  \\
		10 & 3.373 & 0.100  \\
		\hline
	\end{tabular}
	\label{tab:timeApproach}
\end{table}

\subsection{Tree Counting}
\label{subsubsec:comparisionApproaches}

To analyse the design of the proposed architecture, we compared it with a baseline model that does not include the PPM and the multi-stage refinement on tree-counting dataset. The overall best result with just the baseline of the CNN was obtained with a $\sigma = 1$, returning an MAE, RMSE, R$^2$, Precision, Recall, and F-Measure equal to 2.85, 3.72, 0.977, 0.814, 0.833 and 0.822, respectively.

\begin{figure*}[ht]
	\centering

	\subfigure{\includegraphics[width=.24\textwidth]{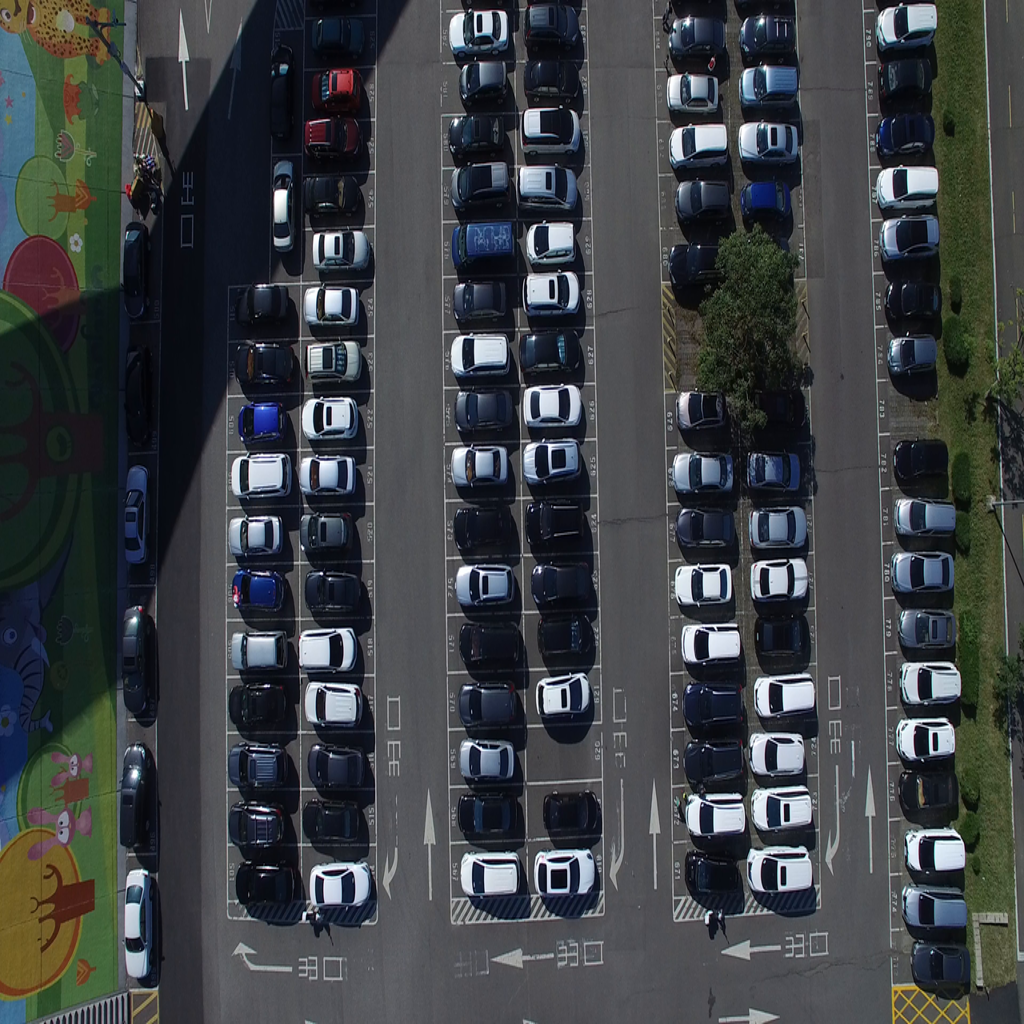}}
	\subfigure{\includegraphics[width=.24\textwidth]{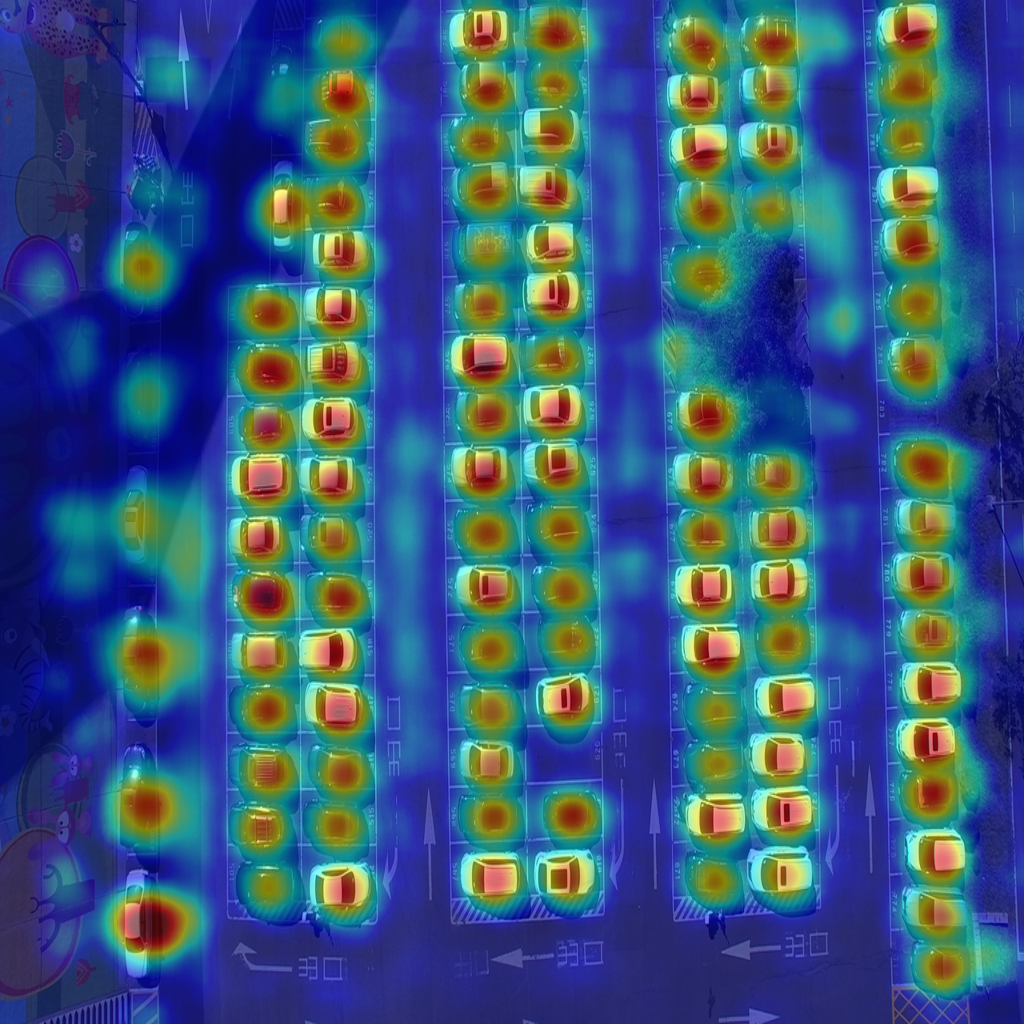}}
	\subfigure{\includegraphics[width=.24\textwidth]{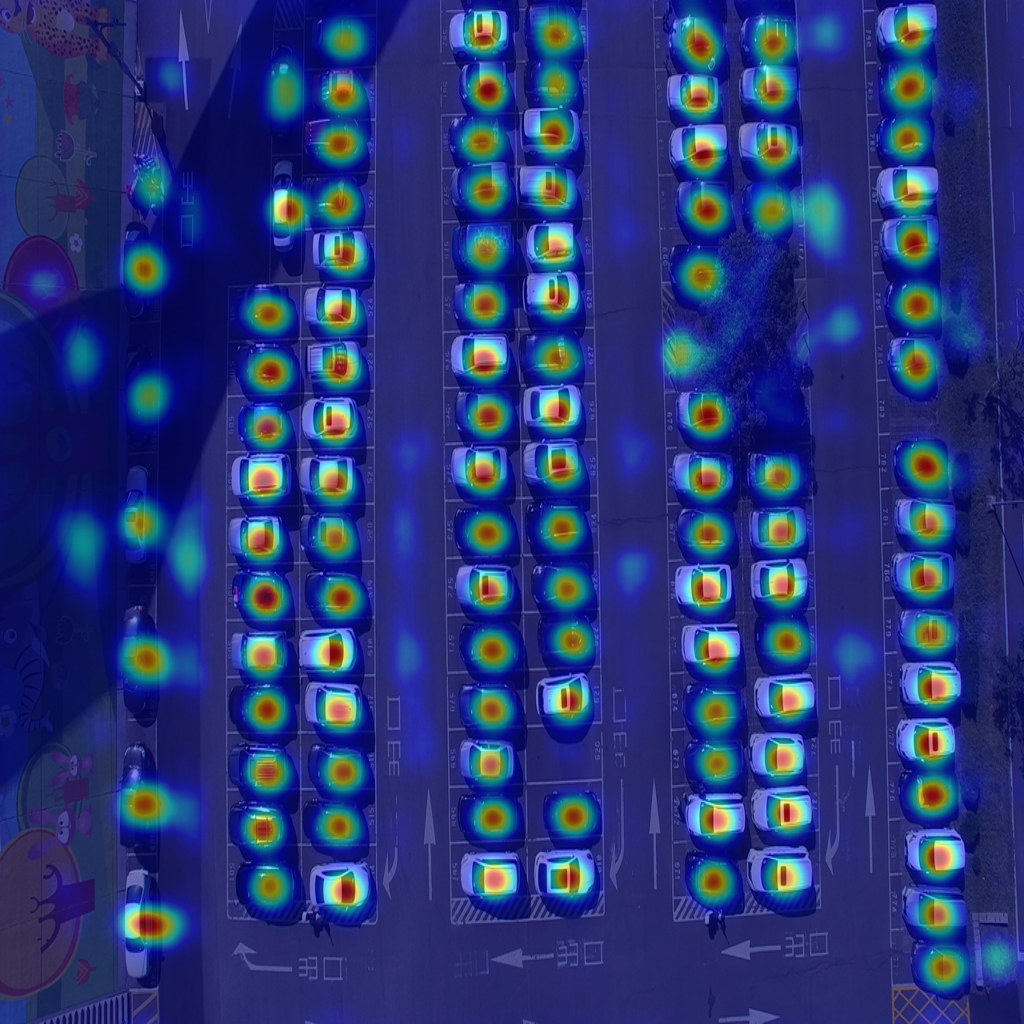}}
	\subfigure{\includegraphics[width=.24\textwidth]{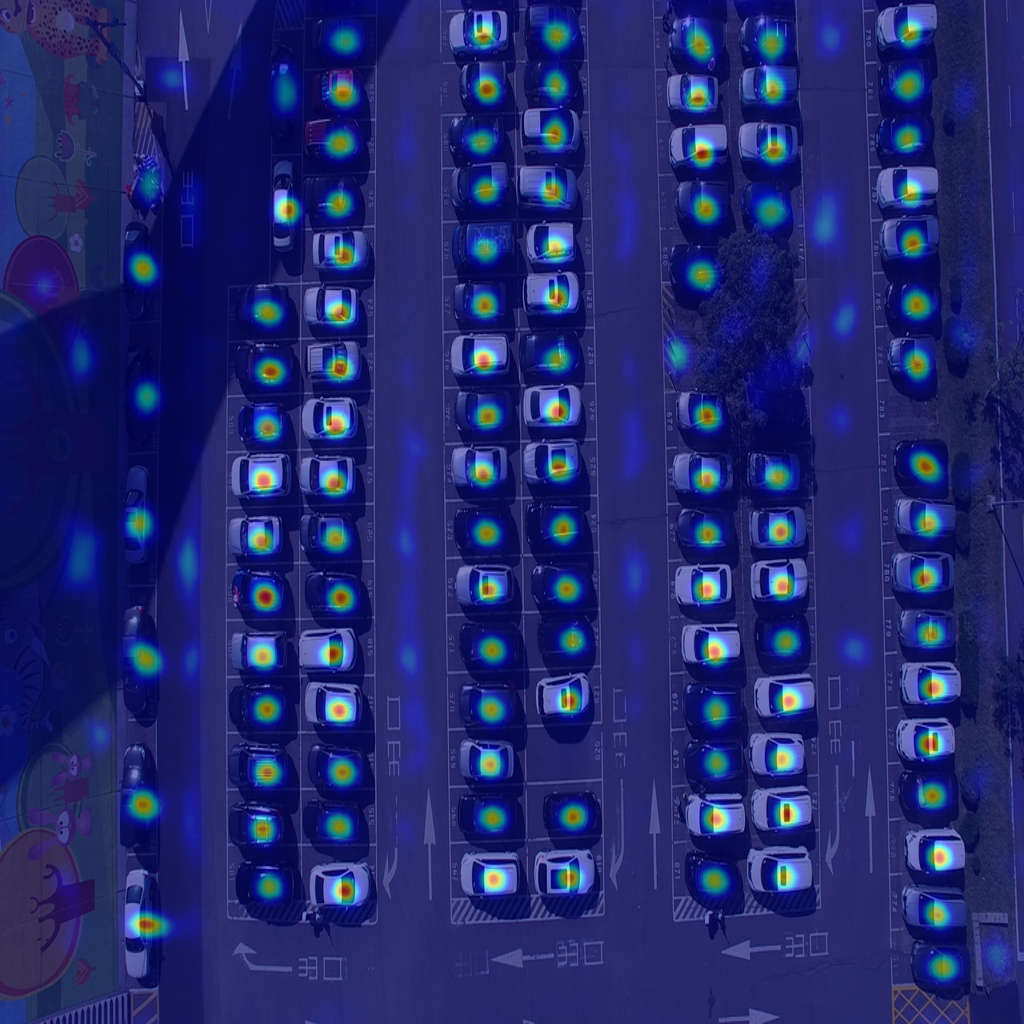}}
	
	\setcounter{subfigure}{0}
	\subfigure[RGB Image]{\includegraphics[width=.24\textwidth]{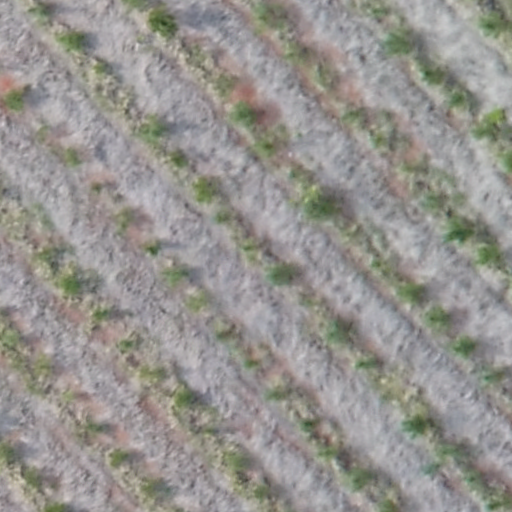}}
	\subfigure[Stage 1]{\includegraphics[width=.24\textwidth]{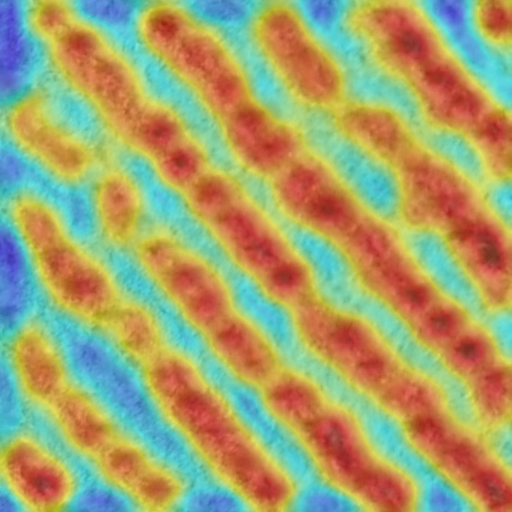}}
	\subfigure[Stage 3]{\includegraphics[width=.24\textwidth]{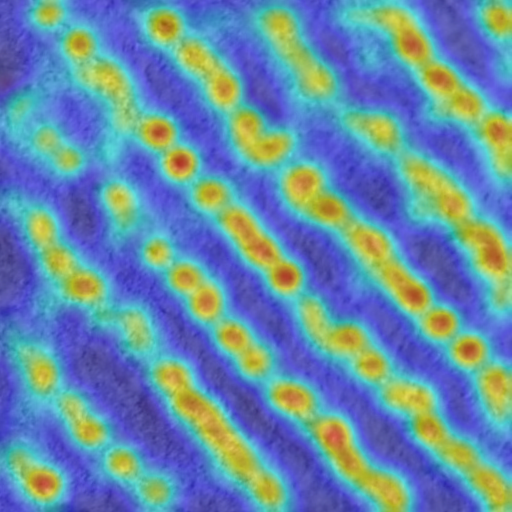}}
	\subfigure[Stage 4]{\includegraphics[width=.24\textwidth]{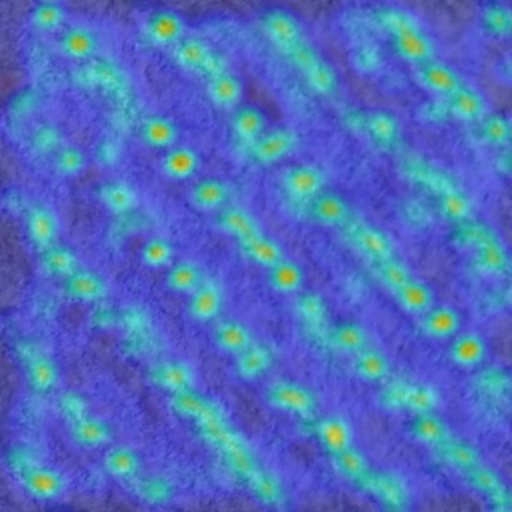}}
	
	\caption{\label{fig:activationStages} Example of two images showing the confidence map refinement by our method.}
\end{figure*}

A gain in performance is observable when analyzing the results from the inclusion of the PPM and multi-stage refinement in the baseline (Table \ref{tab:results}). The inclusion of the PPM has no significant improvement for the results, while the baseline with multi-stage refinement achieves better results. One explanation for this is that multiple stages provide hierarchical learning of the object position, starting from a rough to a more refined prediction of the center of the object. Examples of the confidence map refinement across the stages are shown in Fig. \ref{fig:activationStages}. Besides, when we implemented both these two modules, it outperformed all the baseline results. This shows that the combination of these two modules is essential to object counting.

\begin{table}[ht]
	\caption{Results of the proposed method and its baseline for the tree couting dataset.}
	\centering
	\footnotesize
	\begin{tabular}{|c||c||c||c||c||c||c|}
		\hline
		Method & MAE & RMSE & R$^2$ & Precision & Recall & F-Measure \\
		\hline
		Baseline ($\sigma$ = 0.5)  & 11.97 & 15.10 & 0.62 & 0.861 & 0.709 & 0.772 \\
		Baseline ($\sigma$ = 1.0) & 2.85 & 3.72 & 0.977 & 0.814 & 0.833 & 0.822 \\
		Baseline ($\sigma$ = 2.0) & 3.07 & 4.37 & 0.968 & 0.822 & 0.805 & 0.812 \\
		\hline
		Baseline + PPM & 2.44 & 3.38 & 0.981 & 0.825 & 0.836 & 0.829 \\
		\hline
		Baseline + multi-stage & 2.78 & 3.64 & 0.978 & 0.808 & 0.833 & 0.819 \\
		\hline
		\textbf{Proposed Method} & \textbf{2.05} & \textbf{2.87} & \textbf{0.986} & \textbf{0.822} & \textbf{0.834} & \textbf{0.827} \\
		\hline
	\end{tabular}
	\label{tab:results}
\end{table}

\begin{figure}[ht]
	\centering
	\subfigure[Planting Lines]{\includegraphics[width=.48\columnwidth]{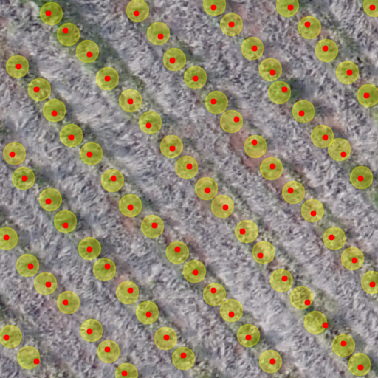}}
	\subfigure[Non-regular Planting]{\includegraphics[width=.48\columnwidth]{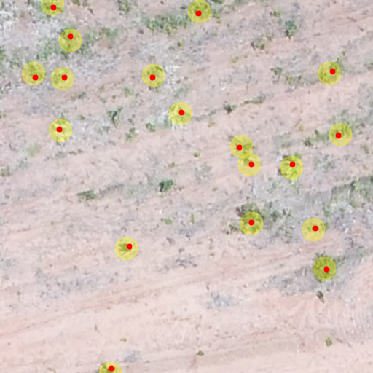}}
	\caption{Comparison of predicted positions (red dots) in two images with different tree density.}
	\label{fig:melhoresResultados}
\end{figure}

We considered a region around the labeled object position to analyze qualitatively the proximity of the prediction to the center of the object. The results using the best configuration ($\sigma_{min}$ = 1.0, $\sigma_{max}$ = 3.0, and $T$ = 4) are displayed in Fig. \ref{fig:melhoresResultados}.
The predicted positions are represented by red dots, and the tree-canopy regions are represented by yellow circles whose center is the labeled position. The proposed method can correctly predict most of the tree positions. Another important contribution is that planting-lines are also identified without the need for annotation or additional procedures (Fig. \ref{fig:melhoresResultados} (a)). Furthermore, the proposed method can correctly identify trees even outside the planting lines, in a non-regular distribution (Fig. \ref{fig:melhoresResultados} (b)).

\begin{figure*}[ht]
	\centering
	\subfigure{\includegraphics[width=.3\textwidth]{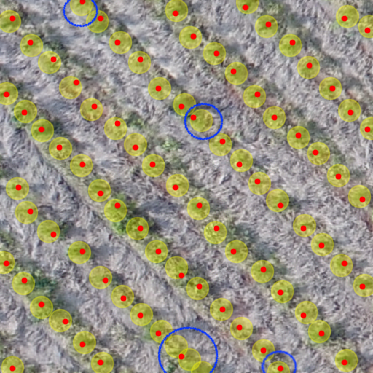}}
	\setcounter{subfigure}{0}
	\subfigure[Proposed Method]{\includegraphics[width=.3\textwidth]{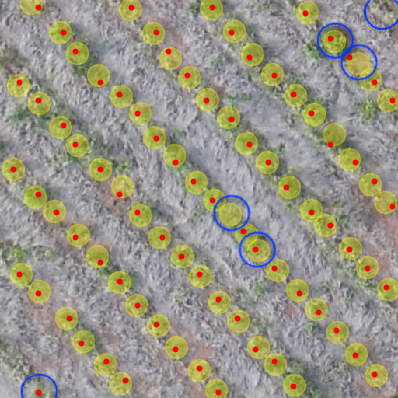}}
	\subfigure{\includegraphics[width=.3\textwidth]{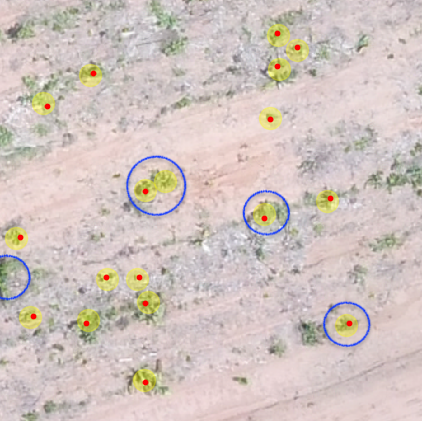}}\\
	\subfigure{\includegraphics[width=.3\textwidth]{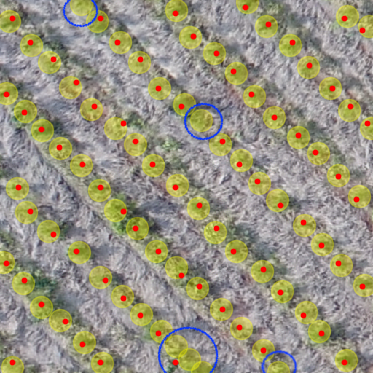}}
	\setcounter{subfigure}{1}
	\subfigure[Baseline]{\includegraphics[width=.3\textwidth]{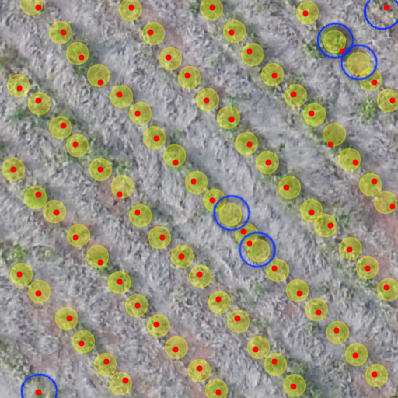}}
	\subfigure{\includegraphics[width=.3\textwidth]{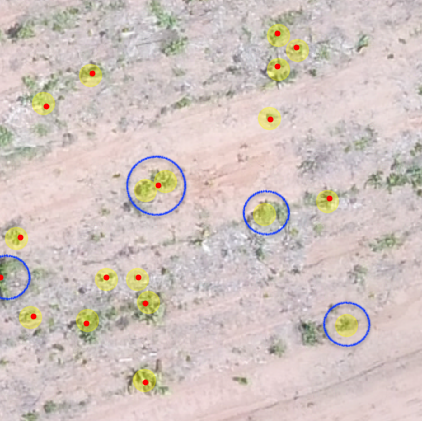}}
	
	\caption{Comparison of the predicted positions of (a) the proposed method and (b) the baseline. Predicted positions are shown by red dots while tree-canopies are represented by yellow circles. Blue circles show the challenges faced by the methods.}
	\label{fig:comparacaoVisual}
\end{figure*}

A comparison of the proposed method with both PPM and multi-stage refinement on the baseline is displayed in Fig. \ref{fig:comparacaoVisual}. The baseline fails to detect some trees while returning some false-positives. The proposed method is capable of detecting more difficult true-positives, not detected by the baseline methods, with fewer false-negatives.

\begin{figure}[ht]
	\centering
	\setcounter{subfigure}{0}
	\subfigure[Short Planting Lines]{\includegraphics[width=.48\columnwidth]{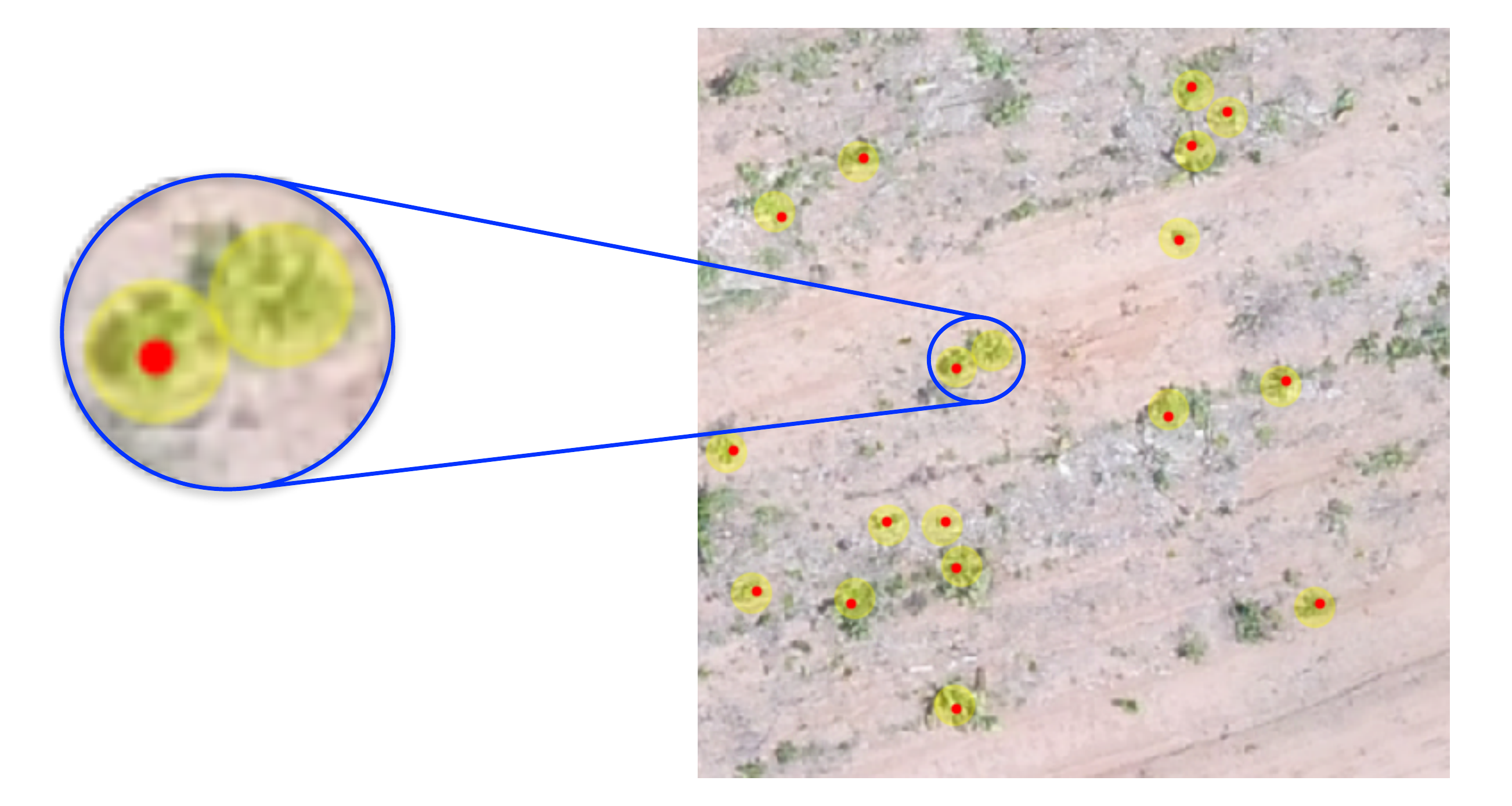}}
	\subfigure[Canopy Occlusion]{\includegraphics[width=.48\columnwidth]{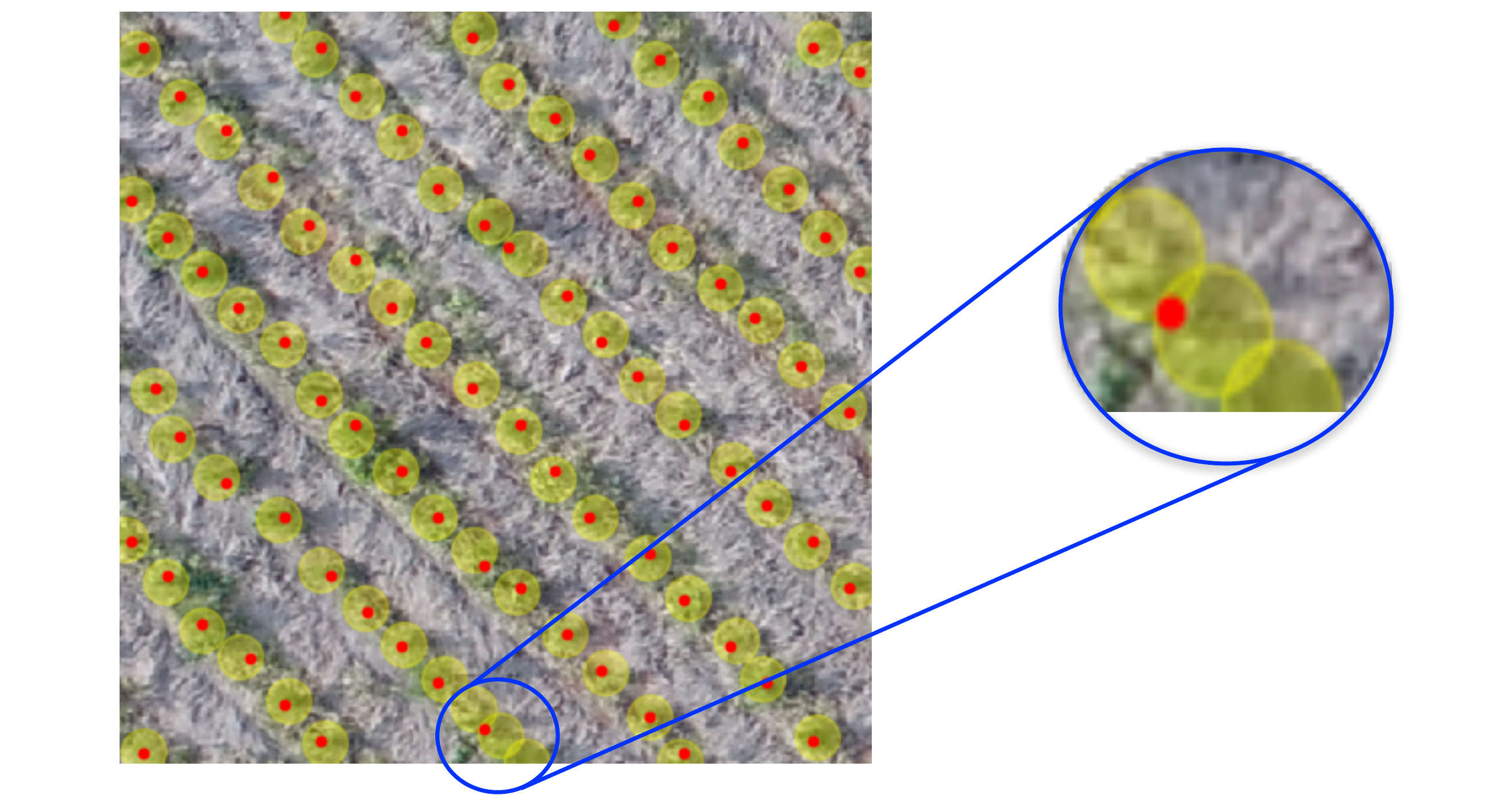}}
	
	\caption{Examples of the challenges faced by the proposed method.}
	\label{fig:desafiosVisual}
\end{figure}

Although the proposed method returned a good performance for the tree counting dataset, it also had some challenges (Fig. \ref{fig:desafiosVisual}). The "far-from-center" predictions occurred in short planting-lines (Fig. \ref{fig:desafiosVisual}  (a)) or in dispersed vegetation. This also happened in highly dense areas (Fig. \ref{fig:desafiosVisual}  (b)), although in fewer occurrences. Still, the proposed method was capable of predicting the correct position of the majority of trees.

\subsection{Density Analysis}
\label{subsec:densityAnalysis}

To verify the performance of the proposed approach for object detection in different types of densities, we divided the tree dataset of $250$ images into three density groups: low, medium and high. For this, the images were ordered according to the number of trees annotated, then the three groups were defined based on the quantities of trees in a balanced way. The low corresponds to images that have up to $52$ plants, the medium between $53$ and $78$ plants, and the high above $78$ plants. Thus, the sets of low, medium and high test images were left with $83$, $90$ and $77$, respectively.

Table \ref{tab:resultsDensities} presents the results obtained by the proposed approach at the three density levels. We can see that the approach does equally well at each density level, obtaining better results at the low level achieving an MAE, RMSE, R$^2$, Precision, Recall, and F-Measure equal to 1.70, 2.34, 0.966, 0.818, 0.846 and 0.829, respectively.

\begin{table}[ht]
	\caption{Results of the proposed method for different object densities.}
	\centering
	\begin{tabular}{|c||c||c||c||c||c||c|}
		\hline
		Density Level & MAE & RMSE & R$^2$ & Precision & Recall & F-Measure \\
		\hline
		Low & 1.70 & 2.34 & 0.966 & 0.818 & 0.846 & 0.829 \\
		\hline
		Medium & 2.10 & 2.85 & 0.865 & 0.824 & 0.829 & 0.826 \\
		\hline
		High  & 2.38 & 3.36 & 0.843 & 0.823 & 0.826 & 0.824 \\
		\hline
	\end{tabular}
	\label{tab:resultsDensities}
\end{table}

\begin{figure}[ht]
	\centering
	
	\subfigure{\includegraphics[width=.32\columnwidth]{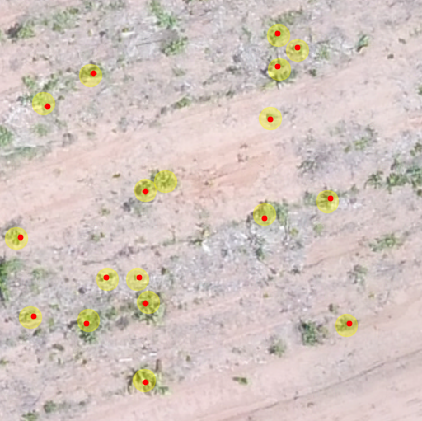}}
	\subfigure{\includegraphics[width=.32\columnwidth]{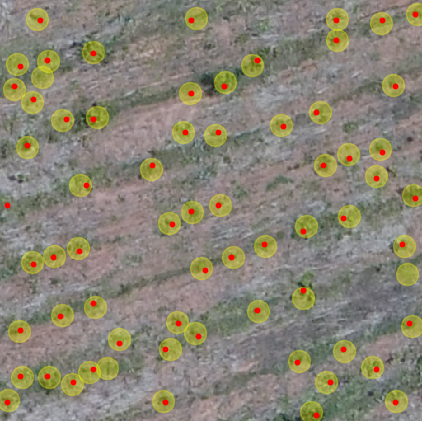}}
	\subfigure{\includegraphics[width=.32\columnwidth]{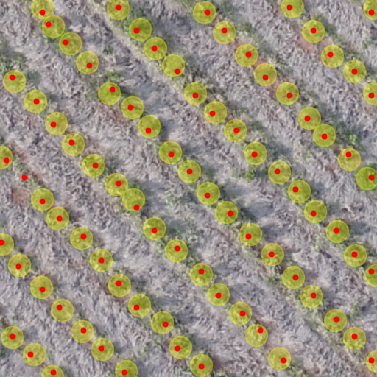}}
	
	\setcounter{subfigure}{0}
	\subfigure[Low]{\includegraphics[width=.32\columnwidth]{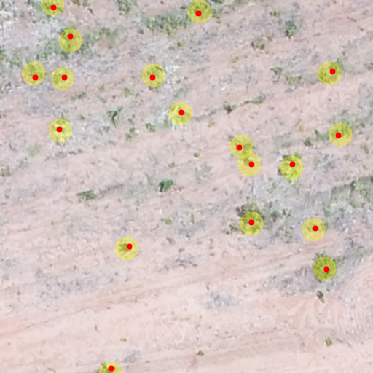}}
	\subfigure[Medium]{\includegraphics[width=.32\columnwidth]{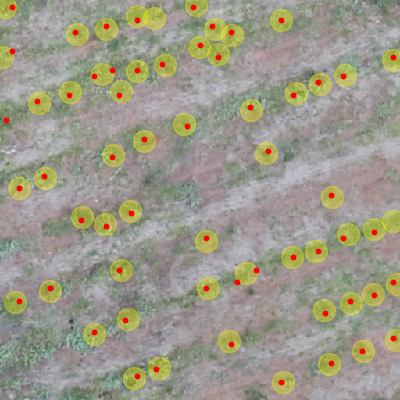}}
	\subfigure[High]{\includegraphics[width=.32\columnwidth]{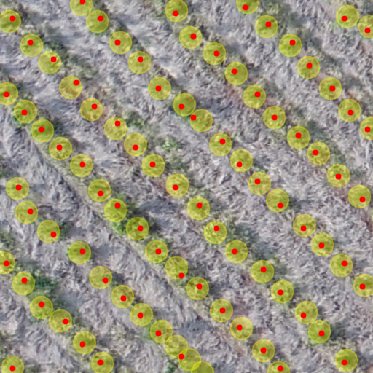}}
	
	\caption{Examples of the performance of the proposed approach at different levels of object densities. Column (a) shows the results for low densities, (b) for medium densities and (c) for high densities.}
	\label{fig:diferentDensities}
\end{figure}

Figure \ref{fig:diferentDensities} shows the visual results for plant detection at the three density levels. We can see that the proposed approach is able to correctly detect the centers of the plants, even in irregular plantings (see Figure \ref{fig:diferentDensities} (a) and (b)).
In addition, as shown in Table \ref{tab:resultsDensities} we can see that at the low level the approach detects the plants positions more easily, since there is not much overlap of the tree canopies.

\subsection{Experiments with CARPK and PUCPR+}

To generalize the proposed approach while comparing its robustness against other state-of-the-art methods, we evaluated its performance on two well-known benchmarks: CARPK and PUCPR+  \citep{Hsieh2017}. These benchmarks provide a large-scale aerial dataset for counting cars in parking lots. The CARPK dataset  \citep{Hsieh2017} is composed of 989 training images (42,274 cars) and 459 test images (47,500 cars). The number of cars per image ranges from 1 to 87 in training images, and from 2 to 188 in test images. PUCPR+  \citep{Hsieh2017} is a subset of the PUCPR dataset \citep{DEALMEIDA20154937}, and it is composed of 100 training images and 25 test images. The training and test images contain respectively 12,995 and 3,920 car instances.

The results from the proposed method and the state-of-the-art methods in both benchmarks demonstrate how feasible our approach is for counting objects (Tables \ref{tab:results_carpk} and \ref{tab:results_pucpr}). We adopted the same protocols for the training and testing sets. The images have been resized to $1024 \times 1024$ pixels since we obtained similar performance when using full-resolution images in our approach. The proposed method achieved state-of-the-art performance in counting objects. Additionally, our method also provides the object position; something not accomplished by most of the high-performance methods. The location of objects can be achieved with traditional object detection methods like Faster R-CNN, YOLO and RetinaNet, which returned inferior results, not being suitable for high object density.

\begin{table}[ht]
	\renewcommand{\arraystretch}{1.3}
	\caption{CARPK Comparative Results.}
	\label{tab:results_carpk}
	\centering
	\footnotesize
	\begin{tabular}{|c||c||c||c||c||c||c|}
		\hline
		Method & MAE & RMSE & R$^2$	& Precision & Recall & F-Measure \\
		\hline
		One-Look Regression \citep{Mundhenk2016} & 59.46 & 66.84 & - & - & -  & -   \\
		IEP Counting \citep{Stahl2019} & 51.83 & - & - & - & -  & -  \\
		YOLO \citep{Redmon2017} & 48.89 & 57.55 & - & - & -  & -  \\
		YOLO9000 \citep{Redmon2017} & 45.36 & 52.02 & - & - & -  & -  \\
		Faster R-CNN \citep{Ren2017} & 24.32 & 37.62 & - & - & -  & -  \\
		RetinaNet \citep{Lin2020,Hsieh2017} & 16.62 & 22.30 & - & - & -  & -  \\
		LPN \citep{Hsieh2017} & 13.72 & 21.77 & - & -  & -  & -  \\
		VGG-GAP \citep{Aich2018ImprovingOC} & 10.33 & 12.89 & - & -  & -  & -  \\
		VGG-GAP-HR \citep{Aich2018ImprovingOC} & 7.88 & 9.30 & - & -  & -  & -  \\
		Deep IoU CNN \citep{Goldman2019CVPR} & 6.77 & 8.52 & - & - & -  & -  \\
		Proposed Method & 4.45 & 6.18 & 0.975 & 0.767 & 0.765 & 0.763 \\
		\hline
	\end{tabular}
\end{table}

\begin{table}[ht]
	\renewcommand{\arraystretch}{1.3}
	\caption{PUCPR+ comparative results.}
	\label{tab:results_pucpr}
	\footnotesize
	\centering
	\begin{tabular}{|c||c||c||c||c||c||c|}
		\hline
		Method & MAE & RMSE & R$^2$	& Precision & Recall & F-Measure \\
		\hline
		YOLO \citep{Redmon2017} & 156.00 & 200.42 & - & - & -  & -  \\
		YOLO9000 \citep{Redmon2017} & 130.40 & 172.46 & - & - & -  & -  \\
		Faster R-CNN \citep{Ren2017} & 39.88 & 47.67 & - & - & -  & -  \\
		RetinaNet \citep{Lin2020,Hsieh2017} & 24.58 & 33.12 & - & - & -  & -  \\
		One-Look Regression \citep{Mundhenk2016} & 21.88 & 36.73 & - & -  & -  & -  \\
		IEP Counting \citep{Stahl2019} & 15.17 & - & - & - & -  & -  \\
		VGG-GAP \citep{Aich2018ImprovingOC} & 8.24 & 11.38 & - & - & -  & -   \\
		LPN \citep{Hsieh2017} & 8.04 & 12.06 & - & - & - & - \\
		Deep IoU CNN \citep{Goldman2019CVPR} & 7.16 & 12.00 & - & - & -  & -  \\
		VGG-GAP-HR \citep{Aich2018ImprovingOC} & 5.24 & 6.67 & - & -  & -  & -  \\
		Proposed Method & 3.16 & 4.39 & 0.999 & 0.832 & 0.829 & 0.830  \\
		\hline
	\end{tabular}
\end{table}

\begin{figure}[ht]
	\centering
	\setcounter{subfigure}{0}
	\subfigure[Occlusion by trees and shadows]{\includegraphics[width=.48\columnwidth]{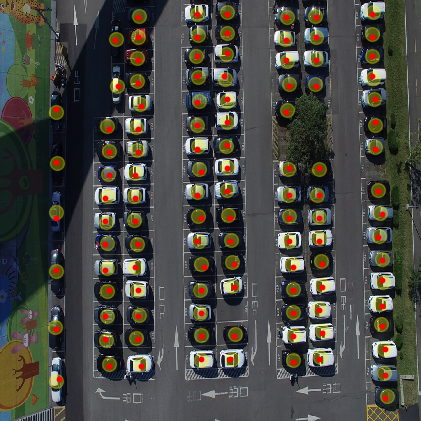}}
	\subfigure[Multiple distances]{\includegraphics[width=.48\columnwidth]{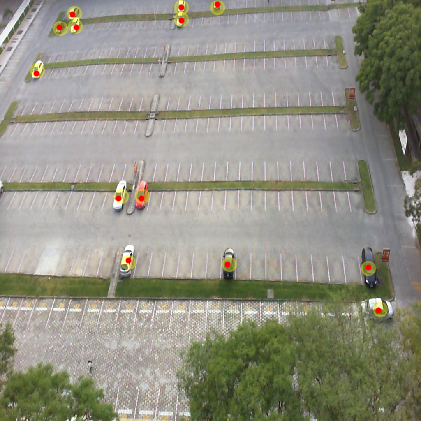}}
	\small
	\caption{Car detection by the proposed method under different conditions.}
	\label{fig:melhoresResultadosCarros}
\end{figure}

As shown in Fig. \ref{fig:melhoresResultadosCarros}, the proposed method improves the results by detecting more difficult true-positives. Some cars are partially covered by trees or shadows while being parked next to each other. Our method was able to detect such cases. The PPM helped improve the object representation, while the multi-stage refinement provided a better position in the center of the objects, especially in highly dense areas. These features, incorporated in our approach, proved to be important additions for both datasets evaluated.

\section{Conclusion}
\label{sec:conclusion}

In this study, we proposed a new method based on a CNN which returned state-of-the-art performance for counting and locating objects with high-density in images. The proposed approach is based on a density estimation map with the confidence that an object occurs in each pixel. For this, our approach produces a feature map generated by a CNN, and then applies an enhancement to the PPM. To improve the prediction of each object, it uses a multi-stage refinement process, and the object position is calculated from the peaks of the refined confidence maps.

Experiments were performed in three datasets with images containing eucalyptus trees and cars.
Despite the challenges, the proposed method obtained better results than previous methods.
Experimental results on CARPK and PUCPR+ indicate that the proposed method improves MAE, e.g., from 6.77 to 4.45 on CARPK and 5.24 to 3.16 on database PUCPR+.
The proposed method is suitable for dealing with high object-density in images, returning a state-of-the-art performance for counting and locating objects. Since this is the first object counting and locating CNN method based on feature map enhancement and a multi-stage refinement of a confidence map, other types of object detection approaches may benefit from the findings presented here.

Further research could be focused on investigating the impact on object counting for different choices of distribution (other than Gaussian) used to generate the groundtruth confidence map.
Predictions other than the confidence map can also help in separating objects of high density, such as predicting the boundaries obtained from the Voronoi diagram.

\section*{Acknowledgments}
This study was supported by the FUNDECT - State of Mato Grosso do Sul Foundation to Support Education, Science and Technology, CAPES - Brazilian Federal Agency for Support and Evaluation of Graduate Education, and CNPq - National Council for Scientific and Technological Development. The Titan V and XP used for this research was donated by the NVIDIA Corporation.

\bibliographystyle{unsrtnat}


\end{document}